\documentclass[3p,times]{elsarticle}

\usepackage{amssymb}

\usepackage{natbib}
\usepackage[figuresright]{rotating}

\makeatletter
\def\ps@pprintTitle{%
  \let\@oddhead\@empty
  \let\@evenhead\@empty
  \let\@oddfoot\@empty
  \let\@evenfoot\@oddfoot
}
\makeatother

\begin{document}
\newtheorem{lemma}{Lemma}
\begin{frontmatter}




\title{New SVD based initialization strategy for Non-negative Matrix Factorization}


\author{Hanli Qiao}
\date{}
\address{Via Carlo Alberto 10, 10123 Torino, Italy \\Department of Mathematics \lq\lq G. Peano\rq\rq, University of Torino}

\begin{abstract}
There are two problems need to be dealt with for Non-negative Matrix Factorization (NMF): choose a suitable rank of the factorization and provide a good initialization method for NMF algorithms. This paper aims to solve these two problems using Singular Value Decomposition (SVD). At first we extract the number of main components as the rank, actually this method is inspired from \cite{Turk91, Turk91_2}. Second, we use the singular value and its vectors to initialize NMF algorithm. In 2008, Boutsidis and Gollopoulos \citep{Boutsidis08} provided the method titled NNDSVD to enhance initialization of NMF algorithms. They extracted the positive section and respective singular triplet information of the unit matrices $\{C^{(j)}\}_{j=1}^k$ which were obtained from singular vector pairs. This strategy aims to use positive section to cope with negative elements of the singular vectors, but in experiments we found that even replacing negative elements by their absolute values could get better results than NNDSVD. Hence, we give another method based SVD to fulfil initialization for NMF algorithms (SVD-NMF). Numerical experiments on two face databases ORL and YALE \cite{ORL, YALE} show that our method is better than NNDSVD.
\end{abstract}


\end{frontmatter}


\section{Introduction} \label{sec:1}
In 1999, Lee and Seung published a paper with the title Learning the parts of objects by
Non-negative Matrix Factorization (NMF) \citep{Lee99}. They analyzed how NMF could learn the parts of objects for facial images and semantic topics. The non-negativity of NMF only permit the additive combination of multiple basis images to present a face. It's compatible with the intuitive notion of combining parts to form a whole. Two different multiple algorithms for NMF were analyzed in \citep{Lee00}. The main idea of NMF is the use of low rank matrices to approximate large dimension data so that reduce the dimension. NMF can also be applied to a lot of other areas: Ding and his colleagues have deeply researched in data clustering and Combinatorial Optimization (see \cite{Ding05,Ding06,Ding08,Ding08_2}). In paper \cite{Yuan05}, Oja and Yuan proposed Projective Non-negative Matrix Factorization (PNMF) for image compression and feature extraction, whereas NMF can be applied to face detection \cite{Chen01}.

NMF framework can be described as follows: given an original non-negative matrix $Z\in \mathbb{R}_+^{m \times n}$, find $W \in \mathbb{R}_+^{m \times p}$ and $H \in \mathbb{R}_+^{p \times n}$, such that: 
\begin{center}
$Z \approx WH$,
\end{center}
where $W$ is called basis matrix, $H$ is called coefficient matrix and $p$ is the rank of the factorization. Note that it is an important index to determine the size of these two low rank matrices. To reduce the dimension of $Z$, we want $p$ is small. On the other hand, for the accuracy of the approximation, the larger $p$ is, the more accurate the approximation will be. Almost all of the researchers set $p$ as different numbers for the first step. Hence we need to find a method to choose $p$, which is much smaller than $\min \{m,n \}$. Generally, it should satisfy the basic rule $(m+n)p<mn$ \citep{Lee00}. In Subsection \ref{sec:2.1}, we will introduce a method based on SVD to choose $p$ that is called Choosing Rule. Here, we give a brief introduction of SVD.

Given a matrix 
$M \in \mathbb{F}^{m \times n}$, which can be a real or complex matrix, there exists a factorization of the form
M = $U \Sigma V^{*}$,
where $U$ is an $m \times m$ unitary matrix over $\mathbb{F}$, $\Sigma$ is an $m \times n$ diagonal matrix with non-negative real numbers on the diagonal, and the $n \times n$ unitary matrix $V^{*}$ denotes the conjugate transpose of the $n \times n$ unitary matrix $V$. Such a factorization is called a SVD of $M$. 

The diagonal entries of $\Sigma$ are known as the singular values of $M$, which are in descending order. In many cases, the top 10\%, even less of the singular values account for over 90\% of all singular values. This means that we can use $p$, which account for over 90\% information of all  singular values to be as the rank of the factorization. $p$ should satisfy the basic rule 
\begin{eqnarray}\label{star}
(m+n)p<mn
\end{eqnarray}

In papers \cite{,Lee99, Lee00}, Lee and Seung gave us two cost functions to describe NMF problem. One of them is $f(W,H):=\frac{1}{2}\Vert Z-WH \Vert_F^2$ and it can be viewed as the following bound optimization problem: Given non-negative matrix $Z$, find $W$ and $H$ which solve
\begin{eqnarray}\label{Fun1}  
min f(W,H), \hspace{0.3cm} W\ge0,\hspace{0.3cm}H\ge0 
\end{eqnarray}
They used multiplicative update rules and additive update rules to solve this problem. We call them MM and AD algorithm respectively. But in MM and AD algorithms, rank $p$ is set by researchers arbitrarily at the beginning of the algorithms, and the initialization values $W_0^{m \times p}$ and $H_0^{p \times n}$ are chosen randomly. NMF can suffer from  slow convergence, then the whole computational process can become much expensive. Hence we should find good initialization method to make algorithms be more effective. Currently there are some literatures which propose different methods to improve the initialization step of NMF algorithms (see \citep{Boutsidis08, Wild03, Wild04, Amy14, Janecek11}). Papers \citep{Wild03},\cite{Wild04} use Spherical K-Means clustering to produce a structured initialization for NMF. Although this method is effective, it increases the computational complexity. In paper \cite{Amy14}, authors compared six initialization procedures on their Alternating Least Squares (ALS) algorithms, whereas paper \cite{Janecek11} applies population based algorithm to NMF. They used five population based algorithms to compute optimal starting points for single rows of $W$ and single columns of $H$. This kind of method obviously makes the computation at cost more expensive. NNDSVD from paper \citep{Boutsidis08}, of all these initialization methods which we mentioned, is the best of effectiveness and low computational cost. But, when NNDSVD deals with the negative elements of the singular vector, they form the unit rank matrix through singular triplets of $Z$, then extracting positive section and respective singular triplet information and using them to initialize $W$ and $H$. Through a large number of experiments, we find that even we set the negative entries of left and right singular vectors are the absolute values, then we use NNDSVD but only once SVD to initialize $W$, $H$, the results are better than initial NNDSVD. This outcome inspires us to find another rule to improve the initialization. We start with SVD and then use singular vectors and values to initialise $W$, $H$. We call this method SVD-NMF. In SVD-NMF, we only use SVD once.
 
As Boutsidis and Gallopoulos referred in \cite{Boutsidis08}, good initialization strategy should satisfy the following conditions: (i) one that leads to rapid error reduction and faster convergence; (ii) one that leads to overall error at convergence. In particular, we point out SVD-NMF has good property under these two conditions.    

We arrange this paper as follows. Section \ref{sec:2} gives two main contents, the first one is the Choosing rule to compute the rank of the factorization and the second one is that introduce the method SVD-NMF. In Section \ref{sec:3}, numerical experiments show the effectiveness of SVD-NMF for two kinds of NMF algorithms.

\section{SVD-NMF for the initialization of NMF} \label{sec:2}
As we mentioned in Section \ref{sec:1}, the rank $p$ of factorization can be calculated by SVD-NMF. For NMF problem, SVD can be expressed as the follows. For any matrix $Z \in \mathbb{R}_+^{m \times n}$, there exists a factorization with the following form:
\begin{equation} \label{SVD}
Z = \mathbf{U} \mathbf{\Sigma} \mathbf{V}^{'}, 
\end{equation}
where the $n \times n$ unitary matrix $V^{'}$ denotes the transpose of the $n \times n$ unitary matrix $V$   and 
\begin{equation}
\Sigma = \left(
\begin{array}{cc}
\mathbf{\Sigma_1} & \mathbf{0}\\
\mathbf{0} &\mathbf{0}\\
\end{array}
\right)
\end{equation}
Here $\mathbf{\Sigma_1}$ = $diag(\sigma_1,\sigma_2,...,\sigma_r)$, and the diagonal entries are sorted in descending order, i.e. $\sigma_1 \ge \sigma_2 \ge ... \ge \sigma_r >0$, $\sigma_i$, $i=1,2,...r$ being the singular values with $r=rank(Z)$. Generally, the sum of $\sigma_i$ (when $i$ is small) accounts for most of all singular values. For NMF, we need to find matrices $W \in \mathbb{R}_+^{m \times p}$ and $H \in \mathbb{R}_+^{p\times n}$ and now we will use this behaviour to choose $p$, which is the rank of factorization for NMF. 

\subsection{Choosing rule} \label{sec:2.1}
From formula (\ref{SVD}), we get the diagonal matrix $\Sigma$. At first, we make the sum of all non-zero diagonal entries for $\Sigma$, that is $\mathbf{sum_r} = \sigma_1+\sigma_2+...+\sigma_r$, and then we choose the number of singular values which accounts for 90\% of all non-zero diagonal entries, that is $\mathbf{sum_p} = \sigma_1+...+\sigma_p$ so we obtain the Choosing rule:  

\begin{eqnarray} \label{p} 
\mathbf{sum_p} / \mathbf{sum_r}<90\% \hspace{0.3cm} and \hspace{0.3cm} \mathbf{sum_{p+1}} / \mathbf{sum_r}\ge 90\%,  
\end{eqnarray}
This is meaningful because the non-zero entries of $\Sigma$ are the square root of non-negative eigenvalues of matrix $ZZ^*$, then we can get that $r \leq \min\{m,n\}$, where $m,n$ are the number of rows and columns of matrix $Z$, respectively. After extracting 90\% energy by the rule (\ref{p}), we can obtain $p \ll r$.
Here we give the MATLAB code of Choosing Rule:
\begin{verbatim}
function [u s v p]=ChoosingR(Z)
[u,s,v] = svd(Z);
sum1= sum(s);
sum2=sum(sum1);
extract=0;
p = 0;
dsum=0;
while(extract/sum2<0.90) 
    p = p + 1;
    dsum=dsum+s(p,p);
    extract=dsum;
end
end
\end{verbatim}

Table \ref{tab1} gives us two groups of rank $p$ for different image matrices using Choosing Rule. These ten images derive from ORL \cite{ORL} and YALE \cite{YALE} database, respectively. ORL face database has 10 different images of each of 40 distinct subjects and the size of each image is $92 \times 112$. YALE face database contains 165 gray-scale images of 15 individuals. There are 11 images per subject and the pixels of each image are $100 \times 100$. We chose 5 images of the first subject on ORL database and another 5 images on YALE database. We can see that both of $p_1$ and $p_2$ satisfy the basic rule (\ref{star}), actually in most cases, $p$ satisfies the basic rule (\ref{star}) under Choosing Rule (\ref{p}). 

\begin{table}[ht]
\caption{Rank $p$ of factorization for different facial images. The first row is 5 images for the first subject on ORL database and the second row is the same as first row on YALE database}
\begin{center}\label{tab1}
\begin{tabular}{cccccc}
\hline
$p_1$ & 35 & 26 & 35 & 34 & 37\\
\hline
$p_2$ & 45 & 47 & 46 & 42 & 45\\ 
\hline
\end{tabular}
\end{center}
\end{table}

\subsection{SVD based initialization: SVD-NMF} \label{sec:2.2}

In this paper, $p$ in numerical experiments are chosen by Choosing Rule (\ref{p}), which is introduced in Subsection \ref{sec:2.1}. As we mentioned good initialization can make convergence fast and get low cost of computational process. NNDSVD uses singular triplets of SVD twice to initialize matrices $W^{m,p}$ and $H^{p,n}$ for NMF. Our new method SVD-NMF only uses SVD once to obtain the singular triplets of $Z^{m,n}$. 

For analyzing the Bound Optimization problem (\ref{Fun1}), we need Eckart–Young Theorem \cite{Higham89} 

\begin{lemma}
Let $V \in \mathbb{R}^{m \times n}$ be a singular value decomposition
\begin{center}
$V = P \Sigma Q^T, \hspace{0.2cm}\Sigma = diag(\sigma_1, \sigma_2,..., \sigma_n) \in \mathbb{R}^{m,n}$,
\end{center}
where $\sigma_1 \ge \sigma_2 \ge ... \ge \sigma_n \ge 0$ are the singular values of $V$ and $P \in \mathbb{R}^{m,m} and Q \in \mathbb{R}^{n,n}$ are orthogonal matrices. Then for $1 \leq r \leq n$, the matrix
\begin{equation} \label{L-F}
B_r = P diag(\sigma_1, \sigma_2,...,\sigma_r, \underbrace{0,...,0}_{n-r})Q^T
\end{equation}
is a global minimizer of the optimization problem
\begin{equation} \label{L-O}
\min \{\hspace{0.1cm}\Vert V-B \Vert_F^2  \hspace{0.3cm} \vert B \in \mathbb{R}^{m,n}, \hspace{0.2cm}  rank(B) \leq r \}
\end{equation}
with the corresponding minimum value $\Sigma_{i=r+1}^n \sigma^2$. Moreover, if $\sigma_r>\sigma_{r+1}$, then $B_r$ is the unique global minimizer.
\end{lemma}

From Lemma 1 we can easily use non-negative matrix $Z^{m,n}$ to convert matrix $V$, then we can get that if there exists a matrix $B_r$ that has the form (\ref{L-F}), the bound optimization problem (\ref{Fun1}) can find the global minimizer $B_r$. We can compute the SVD of non-negative matrix $Z^{m,n}$ and we obtain the singular triplets of $Z^{m,n}$: $U^{m,m}, S^{m,n}, V^{n,n}$. Therefore we want to find a unique matrix $B_r$ such that: $B_r = U diag(\sigma_1, \sigma_2,...,\sigma_r, \underbrace{0,...,0}_{n-r})V^T$.
We can easily verify that:
\begin{equation} \label{Pre-initialization}
B_r = U diag(\sigma_1, \sigma_2,...,\sigma_r, \underbrace{0,...,0}_{n-r})V^T = U^{'} \Sigma^{'}V^T = B_p^{'}
\end{equation}
where 
\begin{equation} \label{Pre-W}
U^{'} = \left(
\begin{array}{cccc}
u_{11} & u_{12} & \cdots & u_{1p} \\
u_{21} & u_{22} & \cdots & u_{2p} \\
\vdots & \vdots  & \ddots  & \vdots \\
u_{m1} & u_{m2} & \cdots & u_{mp} \\
\end{array}
\right)
\end{equation}
and
\begin{equation} \label{Pre-H}
\Sigma^{'} = \left(
\begin{array}{cccccc}
\sigma_{1} & 0 & \cdots & 0 & \cdots & 0 \\
0 & \sigma_{2} & \cdots & 0 & \cdots & 0 \\
\vdots & \vdots  & \ddots  & \vdots & \ddots & \vdots\\
0 & 0 & \cdots & \sigma_p & \cdots & 0 \\
\end{array}
\right),
\end{equation}
$U^{'} \in \mathbb{R}^{m,p}$ and $\Sigma^{'} \in \mathbb{R}^{p,n}$. Then the matrix $B_p^{'}$ is the solution of the bound optimization problem (\ref{Fun1}). We can instantly obtain that $WH = B_p^{'}$ which is the solution of (\ref{Fun1}) that we want to obtain. However, the entries of singular vectors of SVD can be negative, we cannot directly use matrices (\ref{Pre-W}) and (\ref{Pre-H}) to initialize $W$ and $H$. We set the negative elements of matrix $U^{'}$ as the absolute values of themselves, then we get $\vert U^{'} \vert$, where $\vert \mathbf{.} \vert$ indicates that all entries of $U^{'}$ are their absolute values. And we make all the negative entries of $\Sigma^{'}V^T$ as their absolute values. Then we get matrix $\vert \Sigma^{'}V^T \vert$. 

We get the initialization formulas of $W$, $H$:
\begin{center}
$W_0 \hspace{0.2cm}= \hspace{0.2cm} \vert U^{'} \vert$, \hspace{0.3cm} $H_0 \hspace{0.2cm}= \hspace{0.2cm} \vert \Sigma^{'}V^T \vert$
\end{center}
then $W_0 H_0 \approx B_r^{'}$. We use MATLAB to compile SVD-NMF to fulfil the initialization of NMF algorithms. It is the following code:

\begin{verbatim}
m = size(A,1);
[u s v p] = ChoosingR(A);
W = abs(u(:,1:p));
H = abs(s(1:p,:)*v');       
\end{verbatim}

From this code we can see that SVD-NMF only computes the singular triplets of original matrix once, hence it is very simple and it can be used the initialization step for any NMF algorithms. Moreover we can get more stable results of factorization after applying SVD-NMF. For analysing our method SVD-NMF, we combine the initialization step with the algorithms: MM and local non-negative matrix factorization (LNMF) (see \cite{Stan01}). We must note that LNMF is based on another object function Kullback–Leibler divergence $D\hspace{0.1cm} (\hspace{0.1cm} Z\hspace{0.1cm} \Vert \hspace{0.1cm} WH\hspace{0.1cm})$ in \citep{Lee99, Lee00}.

In the paper \cite{Lee00}, the rules of MM algorithm can be expressed as the following form:
\begin{center}
$H \leftarrow H.*((W^T A)./(W^T WH))$\\
$W \leftarrow W.*((AH^T)./(WHH^T))$
\end{center}  
and the updated iterative rules of LNMF in paper \cite{Stan01} has the following form:
\begin{center}
$H \leftarrow \sqrt{ H.*(W^T*( A./(WH)))}$\\
$W \leftarrow W.*((AH^T)./(WHH^T))$

\end{center}  
We also can apply SVD-NMF to other well known NMF algorithms, such as PNMF in paper \cite{Chen01}. In this paper we do not do that. In next Section \ref{sec:3}, we will give some numerical results to show error, the number of iteration and the factorization rank for SVD-NMF, NNDSVD and RnadomNMF, which choose rank $p$ randomly. And the error in this paper we used as the following form:

\begin{center}
\Large{
$\frac{\Vert Z-WH\Vert_F}{\Vert Z \Vert_F}$}
\end{center}

\section{Numerical experiments} \label{sec:3}
In this section, we will show numerical results on two facial databases: ORL and YALE which are already introduced in Subsection \ref{sec:2.1} and two images that are included in MATLAB database. One is football.jpg and another one is kids.tif. The former image has 256 $\times$ 320 pixels, while the latter image has 318 $\times$ 400 pixels. 

\subsection{Numerical results for MM algorithm}
Table \ref{tab2} and Table \ref{tab3} show the errors after using three methods for MM algorithm: SVD-NMF, NNDSVD and RandomNMF. There are 5 images of the first subject on ORL face database and YALE database, respectively, and the number of iterations is 100. We can see that among the three methods SVD-NMF always has the smallest error. This means that SVD-NMF can reach the convergent direction in less iterations than other two methods. We must note that the results of RandomNMF can not be stable. Because the initialization of matrices $W$ and $H$ are chosen randomly, the results can be different at every experiments. Sometimes NNDSVD is worse than RandomNMF. 

\begin{table}[ht]
\caption{The Errors of five image matrices of the first subject on ORL face database by MM algorithms,  the number of iterations is 100.}
\begin{center} \label{tab2}
\begin{tabular}{r||c||c||c||c||c}
\hline
$p$ & 35 & 26 & 35 & 34 & 37 \\ 
\hline
SVD-NMF & 0.1015 & 0.0931 & 0.1039 & 0.1061 & 0.1041\\
\hline
NNDSVD & 0.1149 & 0.0965 & 0.1132 & 0.1084 & 0.1203\\
\hline
RandomNMF & 0.1215 & 0.1098 & 0.1207 & 0.1173 & 0.1178\\
\hline
\end{tabular}
\end{center}
\end{table}

\begin{table}[ht]
\caption{The Errors of five image matrices of the first subject on YALE face database by MM algorithms, the number of iteration is 100.}
\begin{center} \label{tab3}
\begin{tabular}{r||c||c||c||c||c}
\hline
$p$ & 45 & 47 & 46 & 42 & 45 \\ 
\hline
SVD-NMF & 0.1329 & 0.1440 & 0.1500 & 0.1399 & 0.1309\\
\hline
NNDSVD & 0.1319 & 0.1466 & 0.1617 & 0.1460 & 0.1349\\
\hline
RandomNMF & 0.1360 & 0.1419 & 0.1529 & 0.1430 & 0.1345\\
\hline
\end{tabular}
\end{center}
\end{table}

Table \ref{tab4} and \ref{tab5} show the errors of the same image matrices as Table \ref{tab2} and \ref{tab3}. We set iteration number equal to 300. In this case, we can see that SVD-NMF still preserves goodness in having small errors. But there are some exceptions: the second and fourth object in Table \ref{tab4} and the last objects in Table \ref{tab5}. This means that in some cases, NNDSVD has faster convergence than SVD-NMF, but the difference between them is very small. In most cases, SVD-NMF has smaller error and faster convergence than NNDSVD in MM algorithm for NMF. 

\begin{table}[ht]
\caption{Errors of 5 image matrices of the first subject on ORL face database by MM algorithms, the number of iteration is 300.}
\begin{center} \label{tab4}
\begin{tabular}{r||c||c||c||c||c}
\hline
$p$ & 35 & 26 & 35 & 34 & 37 \\ 
\hline
SVD-NMF & 0.0801 & 0.0756 & 0.0819 & 0.0870 & 0.0853\\
\hline
NNDSVD & 0.0899 & 0.0749 & 0.0888 & 0.0868 & 0.0951\\
\hline
RandomNMF & 0.0948 & 0.0712 & 0.0945 & 0.0900 & 0.0918\\
\hline
\end{tabular}
\end{center}
\end{table}

\begin{table}[ht]
\caption{Errors of 5 image matrices of the first subject on YALE face database by MM algorithms, the number of iteration is 300.}
\begin{center} \label{tab5}
\begin{tabular}{r||c||c||c||c||c}
\hline
$p$ & 45 & 47 & 46 & 42 & 45 \\ 
\hline
SVD-NMF & 0.1075 & 0.1234 & 0.1293 & 0.1210 & 0.1116\\
\hline
NNDSVD & 0.1123 & 0.1248 & 0.1404 & 0.1219 & 0.1100\\
\hline
RandomNMF & 0.1104 & 0.1259 & 0.1317 & 0.1219 & 0.1116\\
\hline
\end{tabular}
\end{center}
\end{table}

\begin{figure}
\begin{center}
\includegraphics[width=12cm]{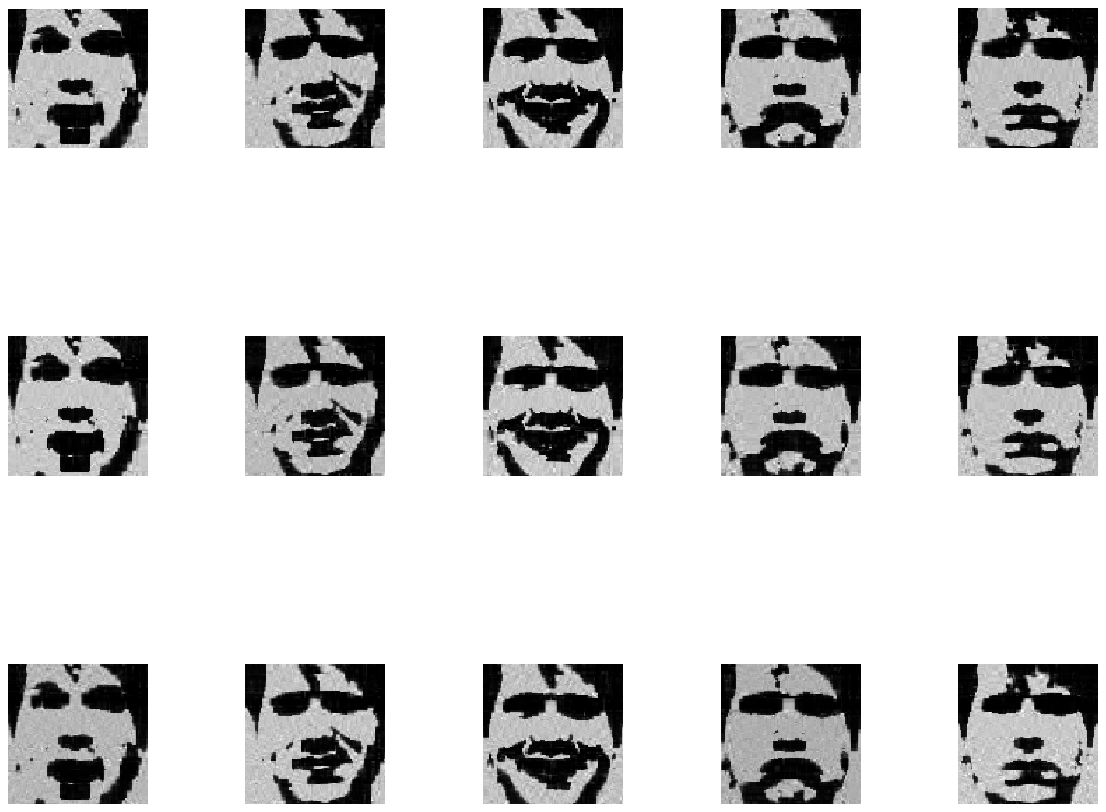}
\centerline{(b) YALE}
\caption{The reconstruction of two face databases. The first row using SVD-NMF, second row using NNDSVD and the last row using RandomNMF for MM algorithm.}
\label{fig_1}
\end{center}
\end{figure}

Figure \ref{fig_1} gives us the reconstruction images using the factorization of SVD-NMF, NNDSVD and RandomNMF for the ten image matrices on ORL and YALE face database, respectively, as we mentioned before. In this case, the iteration number is 100. Because faces on YALE database with more different expressions: happy, sad, normal, sleepy or wink than that of ORL database, therefore the reconstruct results of YALE are a little bit of worse than that of ORL. It deduces that it is more difficult to fulfil face recognition on YALE database than ORL.

\begin{figure}
\begin{center}
\includegraphics[width=12cm]{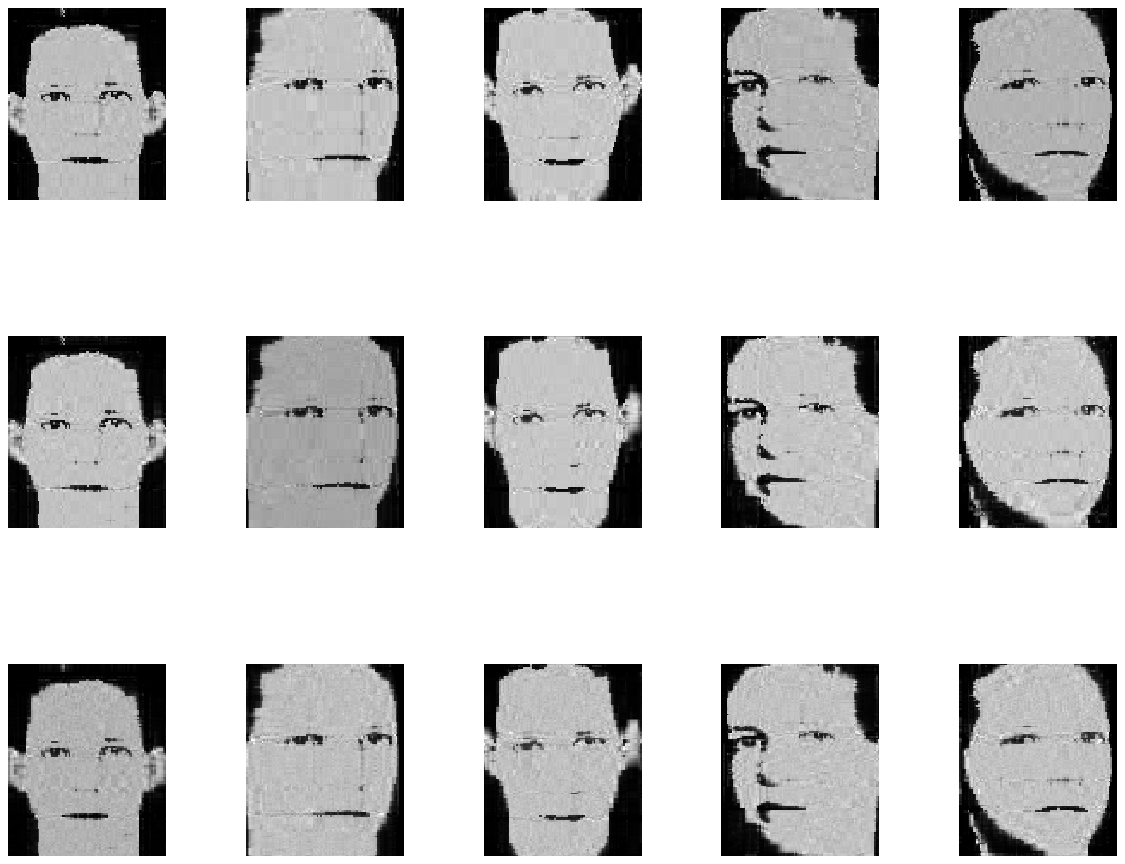}
\centerline{(a) ORL}
\end{center}
\end{figure}

\begin{figure}
\begin{center}
\begin{minipage}{80mm}
\includegraphics[width=8cm]{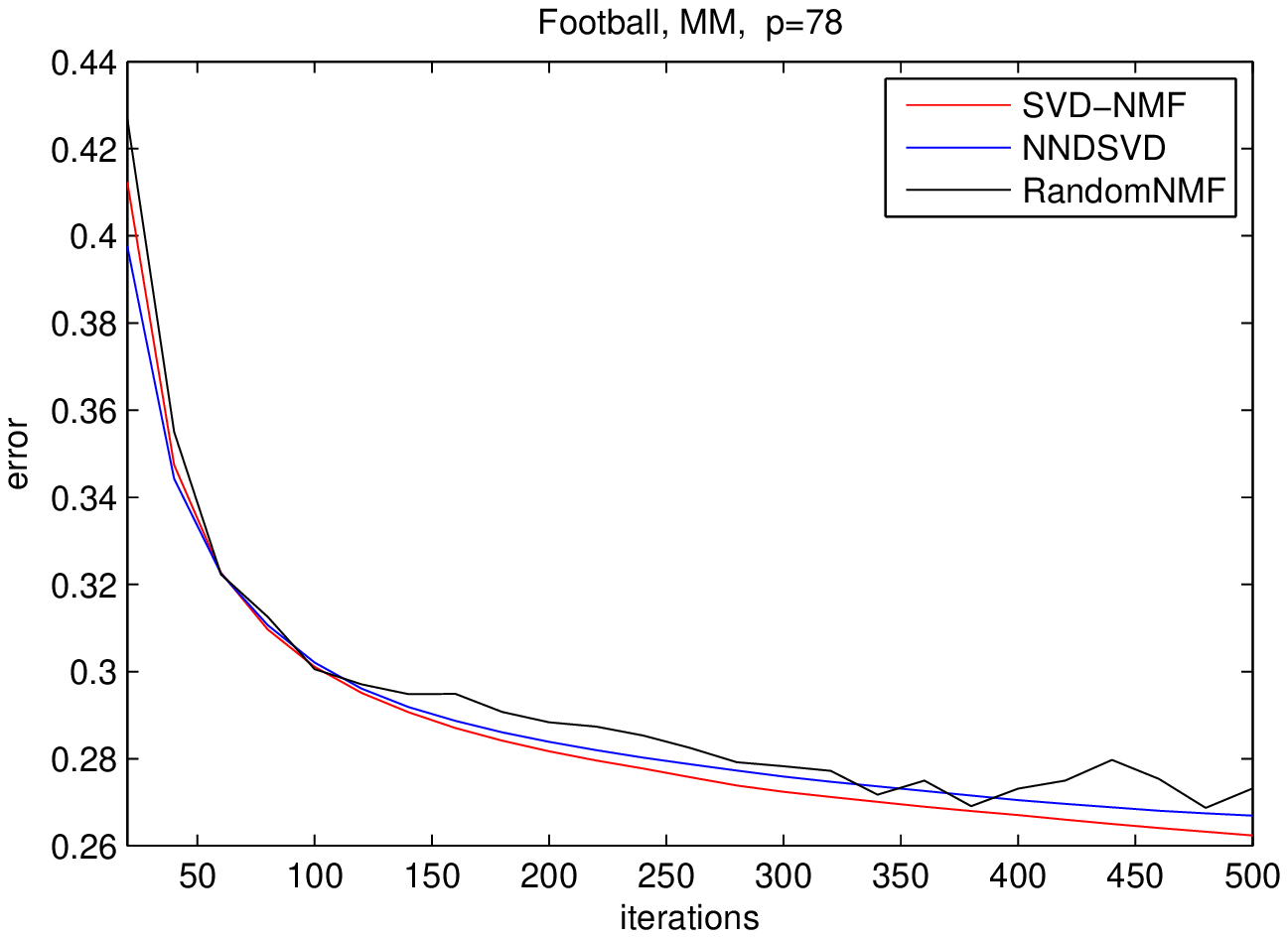}
\end{minipage}
\begin{minipage}{80mm}
\includegraphics[width=8cm]{{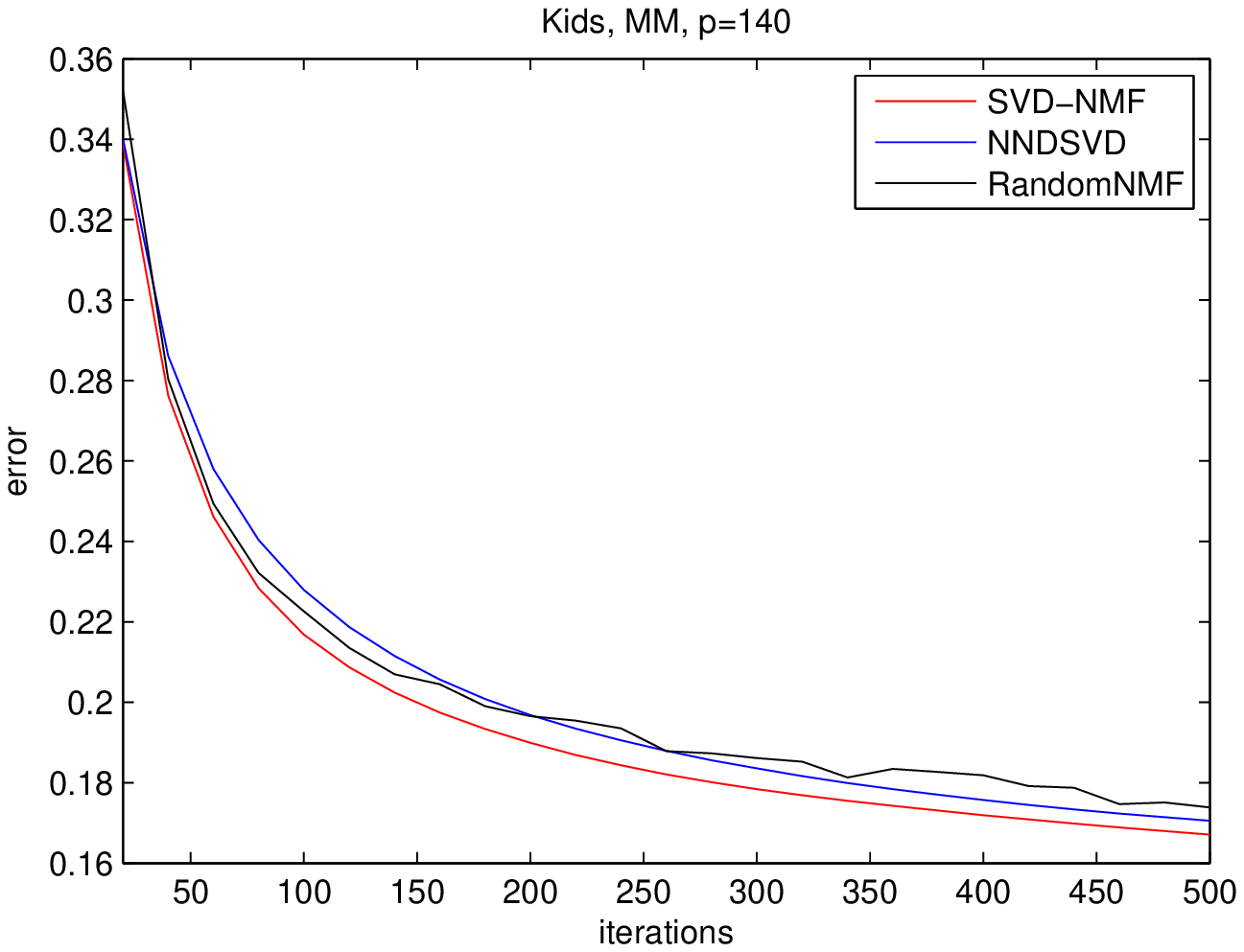}}
\end{minipage}
\caption{The reconstruction of two face databases.}
\label{fig_2}
\end{center}
\end{figure}

Now we use two cases to compare the differences at different iterations of SVD-NMF, NNDSVD and RandomNMF for MM algorithm. The first case is to factorize image football.ipg using these methods. We compute the rank $p$ of factorization is 78. And the second data if from image kids.tif, which the factorization rank $p$ is 140. Figure \ref{fig_2} still shows that SVD-NMF has faster convergence than another two methods in MM algorithm for NMF.

\subsection{Numerical results for LNMF algorithm}
In this section, we will evaluate the properties of SVD-NMF, NNDSVD and RandomNMF methods for LNMF algorithm. Tables \ref{tab6} and \ref{tab7} show the errors of 5 image matrices for second subject on ORL and YALE face database, respectively at 100 iterations. There is the same situation as in Tables \ref{tab2} and \ref{tab3}, that is in most cases SVD-NMF has smaller errors and faster convergence than other methods. After 300 iterations, Tables \ref{tab8} and \ref{tab9} give us the same results. From these four tables, we know that SVD-NMF always preserves fast convergence for LNMF on the second subject image matrices of ORL face database and, in some cases, NNDSVD has faster convergence on YALE database when the number of iterations is relatively large.

\begin{table}[ht]
\caption{Errors of 5 image matrices of the second subject on ORL face database by LNMF algorithms, the number of iterations is 100.}
\begin{center} \label{tab6}
\begin{tabular}{r||c||c||c||c||c}
\hline
$p$ & 51 & 52 & 52 & 52 & 51 \\ 
\hline
SVD-NMF & 0.3955 & 0.3888 & 0.3884 & 0.3905 & 0.3980\\
\hline
NNDSVD & 0.3947 & 0.3947 & 0.3913 & 0.3897 & 0.4009\\
\hline
RandomNMF & 0.4160 & 0.4084 & 0.4030 & 0.4050 & 0.4228\\
\hline
\end{tabular}
\end{center}
\end{table}

\begin{table}[ht]
\caption{Errors of 5 image matrices of the second subject on YALE face database by LNMF algorithms, the number of iteration is 100.}
\begin{center} \label{tab7}
\begin{tabular}{r||c||c||c||c||c}
\hline
$p$ & 42 & 35 & 31 & 35 & 34 \\ 
\hline
SVD-NMF & 0.3934 & 0.4075 & 0.4424 & 0.4140 & 0.4381\\
\hline
NNDSVD & 0.4005 & 0.4064 & 0.4471 & 0.4164 & 0.4438\\
\hline
RandomNMF & 0.4054 & 0.4139 & 0.4472 & 0.4190 & 0.4436\\
\hline
\end{tabular}
\end{center}
\end{table}

\begin{table}[ht]
\caption{Errors of 5 image matrices of the second subject on ORL face database by LNMF algorithms, the number of iteration is 300.}
\begin{center} \label{tab8}
\begin{tabular}{r||c||c||c||c||c}
\hline
$p$ & 51 & 52 & 52 & 52 & 51 \\ 
\hline
SVD-NMF & 0.3571 & 0.3530 & 0.3579 & 0.3557 & 0.3632\\
\hline
NNDSVD & 0.3580 & 0.3547 & 0.3610 & 0.3560 & 0.3672\\
\hline
RandomNMF & 0.3573 & 0.3541 & 0.3603 & 0.3570 & 0.3657\\
\hline
\end{tabular}
\end{center}
\end{table}

\begin{table}[ht]
\caption{Errors of 5 image matrices of the second subject on YALE face database by LNMF algorithms, the number of iteration is 300.}
\begin{center} \label{tab9}
\begin{tabular}{r||c||c||c||c||c}
\hline
$p$ & 42 & 35 & 31 & 35 & 34 \\ 
\hline
SVD-NMF & 0.3761 & 0.4010 & 0.4368 & 0.4031 & 0.4294\\
\hline
NNDSVD & 0.3783 & 0.3994 & 0.4378 & 0.4041 & 0.4287\\
\hline
RandomNMF & 0.3743 & 0.4004 & 0.4375 & 0.4070 & 0.4325\\
\hline
\end{tabular}
\end{center}
\end{table}

\begin{figure}[ht]
\begin{center}
\includegraphics[width=12cm]{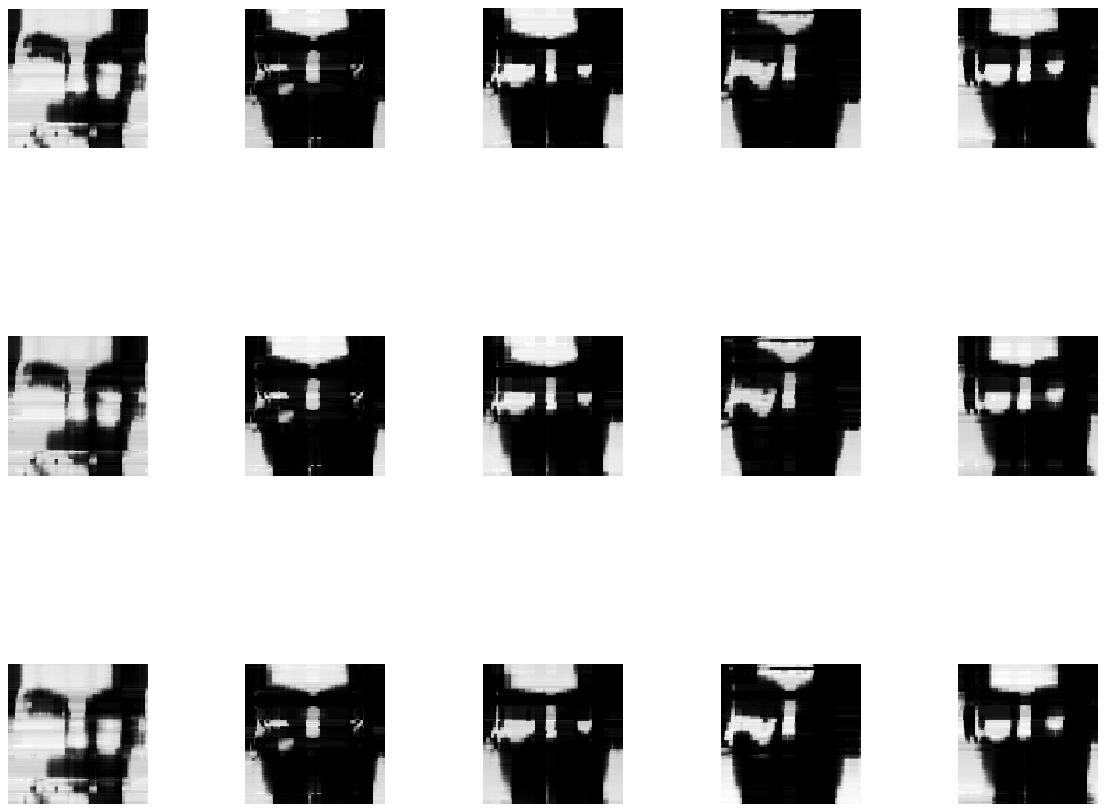}
\centerline{(b) YALE}
\caption{The reconstruction of two face databases and the first row using SVD-NMF, second row using NNDSVD and the last row using RandomNMF for LNMF algorithm.}
\label{fig_3}
\end{center}
\end{figure}

Figure \ref{fig_3} is the reconstruction of the second subjects on ORL and YALE face database. Each subject has 5 images with different expressions, light or other factors. These reconstruct images in Figure \ref{fig_3} have those more details than that of in Figure \ref{fig_1}. This holds because LNMF can impose the local information of the whole face. Images have significant contrasting in (b) of Figure \ref{fig_3} because the first image is affected by lightness whereas others do not have it.

\begin{figure}[ht]
\begin{center}
\includegraphics[width=12cm]{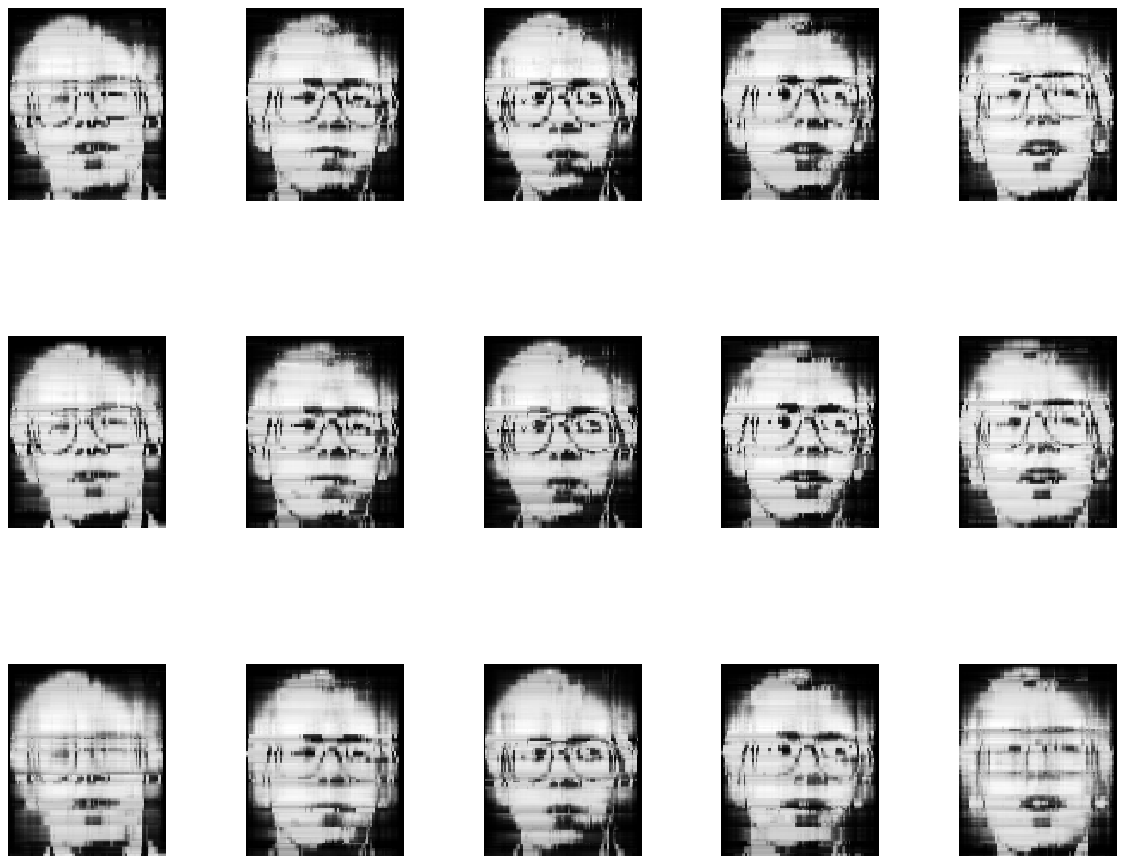}
\centerline{(a) ORL}
\end{center}
\end{figure}

As we introduced in Section \ref{sec:1}, NMF has a lot of applications. Good initialization will bring good factorization, then NMF can be applied well for many fields. SVD-NMF as a new method to impose initialization step of NMF is proposed in this paper. It has some goodness: i) it can be easily combined with other NMF algorithms to enhance the initialization; ii) the computational cost is cheap because we only compute the singular triplets once; iii) it is can reach fast convergence; iv) it is simple and can be satisfied easily. But we still need some other strategy to deal with the negative entries of singular triplets. When we made experiences, we found that if we change the negative elements we can get better or worse results. Hence, how to implement the negative entries rather than replace them by the absolute values is the next step that we should do.

\section{Acknowledgement}

Prof. Alessandra De Rossi and Dr. Roberto Cavoretto who are my tutors, gave me a lot of favours to complete this paper. They gave me a lot of corrections and suggestions. The most important is that they helped me to familiar with the environment of University of Turin so that I can work well in our University. I thank them for their advises and favours very much.





\bibliographystyle{elsarticle-num}
\bibliography{Initialization14_Prepublication}







\end{document}